\documentclass[runningheads]{llncs}
\usepackage[T1]{fontenc}
\usepackage{graphicx}
\usepackage{caption}
\usepackage{subcaption}
\usepackage{wrapfig}
\usepackage[cmex10]{amsmath}
\usepackage {amssymb}
\usepackage{algorithmic}
\usepackage{array}
\usepackage{mdwmath}
\usepackage{mdwtab}
\usepackage{eqparbox}
\usepackage{url}
\usepackage{hyperref}
\usepackage{algorithm}
\usepackage{algorithmic}
\usepackage{epstopdf,cite,color}

\newcommand{\beq}{\begin{equation}}
\newcommand{\eeq}{\end{equation}}
\newcommand{\beqn}{\begin{eqnarray}}
\newcommand{\eeqn}{\end{eqnarray}}
\newcommand{\beqno}{\begin{eqnarray*}}
\newcommand{\eeqno}{\end{eqnarray*}}
\newcommand{\bma}{\begin{displaymath}}
\newcommand{\ema}{\end{displaymath}}
\newcommand{\bnu}{\begin{enumerate}}
\newcommand{\enu}{\end{enumerate}}
\newcommand{\bce}{\begin{center}}
\newcommand{\ece}{\end{center}}
\newcommand{\btb}{\begin{tabular}}
\newcommand{\etb}{\end{tabular}}

\begin{document}
\title{DPFAGA-Dynamic Power Flow Analysis and Fault Characteristics: A Graph Attention Neural Network}
%
%
\author{Tan~Le\inst{1}\orcidID{0000-0003-0807-4357} \and
Van~Le\inst{2}\orcidID{0009-0007-5638-7178} }
\authorrunning{T. Le et al.}
\institute{Hampton University, Hampton, VA 23669, USA\\ 
\email{tan.le@hamptonu.edu}\\
\url{https://sites.google.com/site/thanhtantp} \and
Advanced Technology Center, Virginia Beach, VA 23453, USA.\\
\email{lev13408@gmail.com}}
\maketitle              
\begin{abstract}
We propose the joint graph attention neural network (GAT), clustering with adaptive neighbors (CAN) and probabilistic graphical model for dynamic power flow analysis and fault characteristics. 
In fact, computational efficiency is the main focus to enhance, whilst we ensure the performance accuracy at the accepted level. 
Note that Machine Learning (ML) based schemes have a requirement of sufficient labeled data during training, which is not easily satisfied in practical applications. 
Also, there are unknown data due to new arrived measurements or incompatible smart devices in complex smart grid systems. 
These problems would be resolved by our proposed GAT based framework, which models the label dependency between the network data and learns object representations such that it could achieve the semi-supervised fault diagnosis. 
To create the joint label dependency, we develop the graph construction from the raw acquired signals by using CAN. 
Next, we develop the probabilistic graphical model of Markov random field for graph representation, which supports for the GAT based framework. 
We then evaluate the proposed framework in the use-case application in smart grid and make a fair comparison to the existing methods.

\keywords{Graph Attention Neural Network  \and ML-based Power Prediction \and ML-based Fault Diagnostic Schemes \and Collateral Damage Analysis \and Probabilistic Graphical Model \and  Conditional Random Field.}
\end{abstract}
 \vspace{-0.5cm}
\section{Introduction}
 \vspace{-0.2cm}
Smartgrid is an example of the cyber-physical system that integrates different communications, control and computing technologies to build the massive number of communication and complex networks, which enable to support many critical grid control, monitoring and management operations and emerging applications \cite{tan2016joint,le2015compressed}.
However, it will also be exposed directly to cyber/physical attacks.
Also, there are the raised concerns in the operation and maintenance for smartgrid applications, which are caused by environmental factors, failures of grids/equipments and dynamic variation of demand-response energy.
Here, smartgrid systems usually operate in dynamic environment, whilst the electrical devices have non-linear characteristics and there is the tight relation between the communication and electrical sides.
This is more critical, whenever we consider the large scale network with problems of big data management and complicated correlation.
Moreover, we also need to consider the maintenance and operations, which can address the following incidents, such as disrupted events and load shifting, load balancing, etc.
In short, it is essential to 1) implement the smart monitoring system for acquiring data measurements \cite{tan2016joint, le2015compressed}, 2) derive the AI-empowered framework for analysis, detection and classification of fault and damage \cite{le2022artificial, Zahin20, Zahin20, Tang9461746} and 3) perform avoidance, self-healing and mitigation for collateral damages.
By doing so, we can maximize the quality of the electrical power supply and keep the systems operating efficiently and smoothly.

To tackle these critical challenges and perform the mitigation for collateral damage, we propose the \textbf{Joint Semi-supervised Conditional Random Field and Graph Attention Neural Network for Dynamic Power Flow Analysis and Fault Characteristics}. 
In particular, our main focuses are to develop the frameworks for fault diagnosis, damage analysis and self-healing, which can enhance the resilience and reliability of the powergrid network.
Based on these implemented frameworks, we can utilize them for addressing cybersecurity issues, such as intrusion detection/classification, event-triggered control, etc. 
 \vspace{-0.5cm}
\subsection{Our Proposed Mechanisms}
We aim to perform the fault diagnosis, damage analysis and self-healing for the powergrid system by employing the data-driven intelligent diagnosis methods, which can overcome the critical challenges as presented above. 
In particular, the computational efficiency is the main focus to enhance, whilst we ensure the performance accuracy at the accepted level. 
Note that the ML based schemes require sufficient labeled data during training, i.e. the more labeled data are the higher performance is. 
However, this requirement is not satisfied in the practical application, especially in the large scale network, because we need to obtain the huge dataset as well as there are some new observations/measurements, which are unknown. 
Also, the complexity of smartgrid systems with multiple incompatible smart devices makes us hard to provide enough known labeled data. \\
To respond to this issue, we develop the joint graph attention neural network (GAT), clustering with adaptive neighbors (CAN) and probabilistic graphical model of semi-supervised conditional random field (SSCRF) for fault diagnosis, damage analysis, self-healing and condition monitoring. In fact, the proposed mechanism would model the label dependency between the network data and learn object representations such that it can achieve the semi-supervised fault diagnosis and damage analysis. Here, we develop the graph construction by using the clustering with CAN method to create the joint label dependency. This joint SSCRF and GAT gives reduction of computational latency. We then evaluate the proposed framework in the use-case application of identification of the node status, fault severity and working condition in smartgrid. Based on this analysis and identification, we can implement the intrusion detection and mitigate the damage.

 \vspace{-0.5cm}
\section{Problem Formulation and Graph-Based Architecture Solutions}
 \vspace{-0.2cm}
\subsection{Problem Formulation}

Our main focuses are to develop the frameworks for fault diagnosis, damage analysis and self-healing, which can enhance the resilience and reliability of the power grid network. 
Based on these implemented frameworks, we can utilize them for addressing cybersecurity issues, such as intrusion detection/classification, event-triggered control, etc.

We now present the traditional AC Optimal Powerflow (ACOPF) problem, which is then translated to the machine learning problem \cite{frank2016introduction}. 
Given a grid $\mathcal{G}$, we denote $N$, $L$ and $\mathbb{G}$ ($\mathbb{G} \subseteq \mathbb{N}$) by the set of buses (nodes), the set of branches (edges) and the set of controllable generators, respectively. 
For bus $i$, $P_i^G$, $Q_i^G$, $P_i^L$, $Q_i^L$, $V_i$ and $\delta_i$ are corresponding to the real power injection, the reactive power injection, the real power demand, the reactive power demand, the voltage magnitude and the voltage angle. 
So we formulate the power demand at ACOPF as follows  \cite{frank2016introduction}:
\beqn
\min_{P_i^G} \sum_{i\in\mathbb{G}} C_i(P_i^G) \label{OPT1}\\
\quad \quad s.t. \quad P_i(V,\delta) = P_i^G - P_i^L, \forall i\in\mathbb{N}\\
Q_i(V,\delta) = Q_i^G - Q_i^L , \forall i\in\mathbb{N} \label{OPT2}\\
P_i^G \in [P_i^{G,min}, P_i^{G,max}], \forall i\in\mathbb{G} \label{OPT3}\\
Q_i^G \in [Q_i^{G,min}, Q_i^{G,max}], \forall i\in\mathbb{G} \label{OPT4}\\
V_i \in [V_i^{min}, V_i^{max}], \forall i\in\mathbb{N} \label{OPT5}\\
\delta_i \in [\delta_i^{min}, \delta_i^{max}], \forall i\in\mathbb{N} \label{OPT6}
\eeqn
Here, (\ref{OPT1}) typically represents a polynomial cost function, (\ref{OPT2})--(\ref{OPT3}) correspond to the power flow equations and (\ref{OPT4})--(\ref{OPT6}) represent operational limits on real/reactive power injections, nodal voltage magnitude and nodal voltage angles respectively. 
Note that we also have the constrains for the branch currents (see \cite{frank2016introduction} for  more detail).
This observation only complicate the optimization problem and we can easily address this issue in the extension.
For simplicity, we only consider the simple but remarkably informative problem.
In the following, we translate the original optimization problem of ACOPF into the machine learning problem, which includes 1) End-to-end Prediction and 2) Optimal Constraint Prediction. 
By doing so, we can solve the computationally expensive problem in real-time.
Recall that the real and reactive demands, $P_i^L$ and $Q_i^L$ are assumed to be known for all buses.

\noindent \textbf{End-to-end Prediction:} In machine learning, we observe the ACOPF problem as an alternating regression, in which the pair of demands and responses, ($P_i^L$ and $Q_i^L$) and ($P_i^G$ and $V_i^G$) must be predict efficiently.
After obtaining the results of demands $\mathbf{X} = \{[P_0^L,..., P_N^L, Q_0^L, ..., Q_N^L]\}\mid^n_{i=1}$ and corresponding responses $\mathbf{Y} = \{[P_0^G, ..., P_G^G, V_0^G, ..., V_G^L]\}\mid^n_{i=1}$, we need to train the model of $f_{\theta} : \mathcal{X} \mapsto \mathcal{Y}$ so that the error between the optimal and predicted generator settings ($\mathbf{Y}, Tilde{Y}$) is minimized. 
Finally, we determine the remaining state variables (i.e. solving the power flow problem) and then evaluate $V_i^L$, $Q_i^G$, and $\delta_i$ to satisfy the constraints of (\ref{OPT2})--(\ref{OPT6}).

\noindent \textbf{Optimal Constraint Prediction:} Note that the previous prediction may output the results that violate the constraints.
Hence, we need to develop the optimal constraint prediction, where we learn the possible constraint set that are active for some optimal demand results.
It means that the unknown variables is at the border of the optimization ranges, when the specific constraint is active. 
So, the benefit of this optimal constraint prediction is allowing us to perform a warm start and hence decrease the CPU time.
There are other benefits of the optimal constraint prediction for end-to-end prediction, such as
\textbf{Solver Speedup}, 2) \textbf{Reliability} and 3) \textbf{Task complexity} (see \cite{guha2019machine} for more detailed information). 
In the subsequent sections, we will present the machine learning methods to solve the ACOPF problem.

 \vspace{-0.5cm}
\subsection{Fully Connected Neural Networks}
 \vspace{-0.5cm}
The traditional mechanism using Fully Connected Neural Networks (FCNN) usually works for the case of fully available data. This mechanism works perfectly for the dataset with underlying data representation of grid-like structure. Hence, it is applied to multiple applications, such as AC power flow prediction, image classification and semantic segmentation \cite{Jegou_2017_CVPR_Workshops, Duchesne2020, guha2019machine, Zahin19, Zahin20}. However, it cannot give the same performance, when we apply it to the application with lower train dataset size. In that way, the traditional method needs to acquire more new data, which increase the cost of operations. Otherwise, the performance would be degraded significantly. 
Furthermore, data in many applications, like 3D meshes, social networks, telecommunication networks and power grid networks, would be in the irregular domain and cannot be represented in a grid-like structure.
Also, this method does not provide the accurate performance because it does not consider using of information from neighboring nodes in the power grid network.

\begin{figure}[!t]
\centerline{\includegraphics[width=0.6\textwidth]{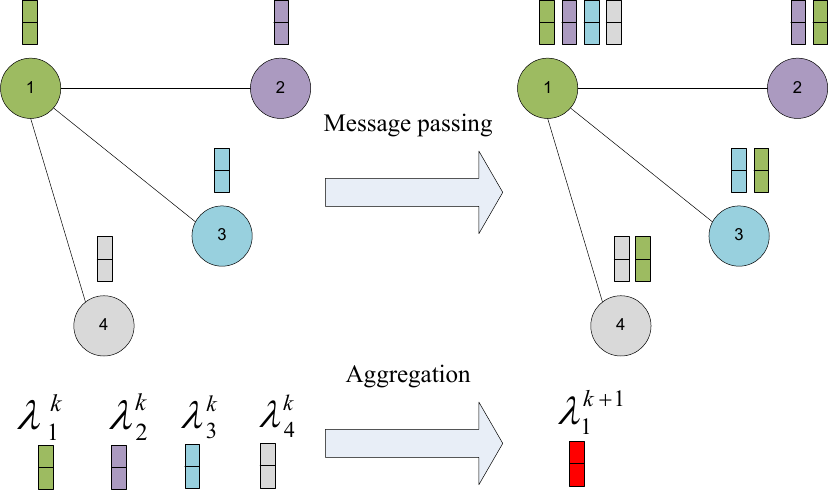}}
	\caption{Message passing operation at node 1 in GNN.}
	\label{messagepass}
 \vspace{-0.5cm}
\end{figure}

So, we recommend using the form of graphs to represent these data and utilize the Graph Neural Network (GNN) to generalize convolutions to the graph domain. The key contribution of GNN is using the message passing with the neighboring nodes to integrate the connected power grid network to the represented graph (see Fig.~\ref{messagepass}). Finally, we suggest using self-attention strategy in addition to the utilization of hidden representations of each node (with the relationship of the neighboring nodes). This method is called the GAT. The attention architecture in GAT can implicitly create the differentiation for the importance levels of different nodes within a neighborhood. This mechanism has a potential to perform node classification and to evaluate the critical nodes of graph-structured data.

\subsection{Graph Neural Networks and Graph Attention Networks}

We firstly propose the GNN and GAT \cite{velickovic2018graph, Qu2019GMNNGM, Tang9461746}, which achieve high expressive capability by assigning adaptive weights to different neighbors. To reduce the complexity, we perform the linearization and approximation for the proposed methods. In the context of cyber resilience, this mechanism can support for big data management because it does not require labels for all the nodes and their data. Moreover, this mechanism fully utilizes the relationship/dependency between network elements so that it can achieve the high efficiency and effectiveness. We will describe the key differences between the traditional method FCNN and the proposed methods of GNN and GAT.

\begin{figure}[!t]
\centerline{\includegraphics[width=0.8\textwidth]{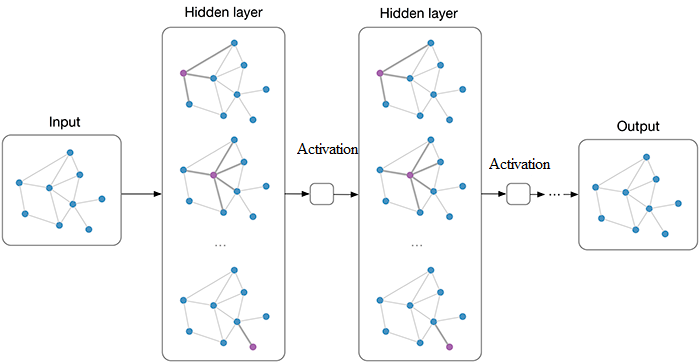}}
	\caption{Structure of GNN.}
	\label{gcn_network}
 \vspace{-0.5cm}
\end{figure}

The first consideration is using GNN, which can extract the locality features and formulate the graph data. Note that the Convolutional Neural Network also has this function, however, it only works on Euclidean data, like text (1D flat shape) and images (2D grid shape). Convolutional Neural Network cannot solve the problems in non-Euclidean space, where the generated data are in graph shape. This is because 1) graph has no fixed node ordering so that it can be served as a reference point; 2) graph has arbitrary size with complex topological structure; 3) there is no spatial locality in the graph.GNN solves this problem by using a different mechanism. GNN allows feature exchange between nodes and its neighbors; Figure 1 demonstrates the message passing operation at node 1, where it receives information from nodes 2, 3 and 4, and then performs the aggregation. GNN's formulated layout is from multiple Convolutional Neural Network layers and multiple non-linear activate functions, such as Sigmoid, Tanh, ReLu, etc, see Fig.~\ref{gcn_network}.

\begin{figure}
    \centering
    \begin{subfigure}{.35\textwidth}
    \centering
    \includegraphics[width=.95\linewidth]{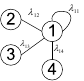}  
    \caption{Message passing at node 1.}
    \label{GAT_messagepassing}
    \end{subfigure}
    \hfill
    \begin{subfigure}{.6\textwidth}
    \centering
    \includegraphics[width=.95\linewidth]{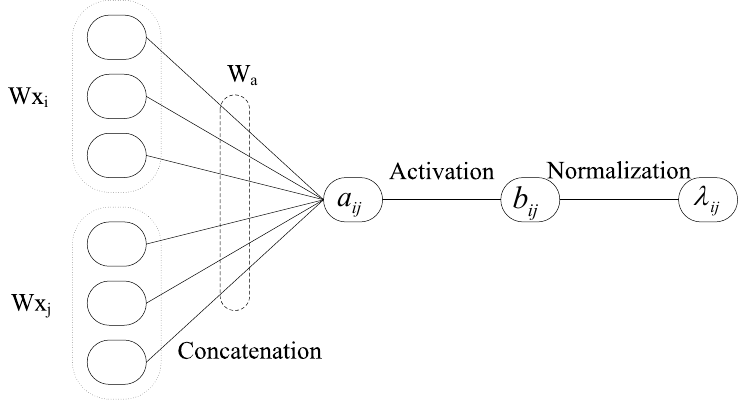}  
    \caption{Summation Message Passing.}
    \label{GAT_summation}
    \end{subfigure}
    \vspace{-0.3cm}
    \caption{Message Passing in GAT.}
    \vspace{-0.3cm}
\end{figure}

As we state above that the key difference between the FCNN and the proposed GNN and GAT is using the message passing, which allows the considered node to collect information from its neighbors. It implies that each node can extract the salient feature from its neighbors. The proposed GAT operates in the similar way. 
For the attention part, it uses the message from the node itself as a query, while it utilizes the passing messages to average both keys and values. It is noted that the operation includes the message to itself as well. For example, Fig.~\ref{GAT_messagepassing} demonstrates the message passing operation at node 1, where it receives information from nodes 2, 3, 4 and itself. Fig.~\ref{GAT_summation} illustrates the sequential operations in GAT, where $a_{ij} = W_a[Wx_i\mid\mid Wxj],$ $b_{ij} = Activation(a_{ij})$ and $\lambda_{ij} = Normalization(b_{ij}).$
Here, activation functions are sigmoid, tanh, relu, etc., whilst normalization function is softmax.
\vspace{-0.3cm}
\subsection{Graph Neural Networks and Graph Attention Networks based Framework for Visualization, Prediction, Detection and Classification}
\label{GNN_GAT_CRF}

In the context of power grids, we need to utilize the available energy data to maximize the opportunities and minimize any possible risks, whenever we perform the management of power system. In particular, we suffer from the increase of energy consumption because we need more energy for heating and transportation. Furthermore, the smart grids involve with the increase of data amount and data flexibility because there are a lot of devices for communication and electrical control. Therefore, there is a need to manage this flexibility to ensure the secure and resilient operations of power systems. One of potential mechanisms is using GNN and GAT to visualize the grid status and connectivity by utilizing the power data (i.e. power and voltage). Specifically, we try to integrate graph topology of the circuit nodes in smart grid systems with GNN architecture. Note that the formulated GNN model will effectively utilize the connectivity data from the graph by involving adjacency matrix. Based on the established graph, we can utilize the features of lines and/or edges in the graph data to support further applications, such as detection and classification of possible risks and attacks as well as abnormal operations. In the following, we present the essential need of our proposed data representation, which is then combined with GAT and/or GNN for training label dependency.

The statistical relation learning methods and GNN methods are often used to learn from graphs.
In this paper, we utilize the probabilistic graphical model of Markov random field in representation part with the purpose of representing the uncertain scenarios and the dependencies within relational data, while we use the GAT in the learning part. 
Note that many innovative statistical modeling methods \cite{lafferty2001conditional,Tanle2024} are applied in pattern recognition and machine learning and are used for structured prediction.
For simplicity, the SSCRF \cite{lafferty2001conditional, Tanle2024, Zahin19, Zahin20} is adopted for probabilistic graphical model, which include the semi-supervised method. 
It is easy to extend to consider the more complex and efficient mechanisms for data representation.
We consider the graph $G$ ($G = (V, E, x_V)$) with the edge set of $E$, the node set of $V$ and node attributes of $x_V$.
We will consider both the labeled nodes $V_L$ and unlabeled nodes $V_U$ in our semi-supervised learning.
So, we denote these corresponding labels by $y_L$ and $y_U$.
Given the edge set of $E$ and node attributes of $x_V$, we will derive the label distribution of $p (y_U|x_V , E)$.
We briefly present the SSCRF and GAT as follows.

\noindent \textbf{1) SSCRF Method}: We propose the SSCRF method to learn the label dependency, which composes of two components, i.e. CRF and semi-supervised learning.
The former of CRF is the probabilistic graphical model of Markov random field and is used for modeling the label dependency.
The later of semi-supervised learning is the hybrid of supervised and unsupervised learning and uses both labeled and unlabeled nodes for training.
It means that our proposed GAT captures dependencies between instance labels.
Note that, we make the differentiation between GAT and GNN.
In the semi-supervised GNN methods, the predictions of network are only determined by the object representations, but the joint dependency of labels is ignored.
We now determine the label distribution of the node set $V$ (where its label matrix denotes by $y_V$).
Let denote $\Phi_{i,j}(y_i, y_j, x_V )$ by the potential function of the connected edge $(i, j)$ in graph $G$,  which aims to create a new feature by taking a weighted sum of different handcrafted features.
We then perform normalization for the result with the normalization factor of $Z(x_V)$. 
In summary, the label distribution is presented as \cite{Tang9461746}
\beqn
p(y_V|x_V , E) = \frac{1}{Z(x_V)} \prod_{(i, j) \in E} \Phi_{i,j}(y_i, y_j, x_V ).
\eeqn

\noindent \textbf{2) GAT Method}: The main different between GAT and GNN is that the GAT employs the attention mechanism each node in the training node representations $h_V$, in which it provides the coefficient $e_{ij}$.
This attention coefficient is calculated by the linear function of the weight matrix for the node set $W h_V$.  
It implies that GAT provides the evaluation of the importance of one node to its neighbors based on the attention coefficient $e_{ij}$
Furthermore, the softmax function is used for normalization, which is denoted by $\alpha_{ij} = soft max(e_{ij})$.
The procedure of aggregation $g(\cdot)$ is employed to update the node representation, i.e. $\tilde{h}_V = g(x_V , E, \alpha_V)$. 
We can observe that the updated node representation is created by multiple relationship factors, i.e. node attributes, edge connections and attention coefficients.
We now can evaluate the label distribution of unlabeled nodes, $p(y_U|x_V, E)$, i.e. $p(y_V|x_V, E) = Cat(y_V|soft max(W \tilde{h}_V ))$, where $Cat(\cdot)$ is the categorical distribution.
We repeat the updating steps to get the satisfied result and then normalize it to obtain the final label distribution.  
So, GAT serves as an end-to-end prediction, where both labeled and unlabeled samples are used for training.

Now, we derive the combination of GAT and CRF with the semi-supervised learning, which has a capable of learning the effective node representations and modeling the label dependency. 
This framework terms as the graph-based semi-supervised intelligent neural network approach, which achieves semi-supervised power flow prediction. 
In particular, our proposed framework is optimized with the variational expectation–maximization algorithm  \cite{velickovic2018graph, Qu2019GMNNGM}, which is composed of expectation stage for inference and maximization stage for learning. 
A GAT \cite{velickovic2018graph, Qu2019GMNNGM} is designed to model label dependency directly, which is used to replace the handcrafted potential functions in SSCRF. 
Recall that we use CAN to translate the raw acquired data to graph to form the graph structure adaptively.

\vspace{-0.5cm}
\subsection{Constructing Graph by CAN}
\label{CAN_TEXT}

We also enhance the accuracy performance by using the CAN method to construct the graph. 
Based on local distances, the CAN method learns connections of a graph by assigning neighbors for each object adaptively \cite{nie2014clustering}. Given $z = \{z_1, z_2, ..., zn\}$, $z \in R_{n \times m}$ is represented as the sample matrix. 
The possibility that the $i$th sample $x_i$ can be the $j$th sample $z_j$ neighbor is $s_{ij}$. 
The main assumption of CAN is that if the distance between samples is smaller, the probability to be neighbors is bigger. 
In the following, we use the traditional Euclidean distance as the distance parameter for simplicity in the presentation of problem.
However, we develop the general algorithms to address both graph distances and traditional Euclidean distances.
Therefore, neighbors are assigned to every sample by solving
the problems in the following equation:
\beqn
\min \sum^{n}_{i, j=1} \parallel z_i - z_j\parallel^2 s_{ij} + \gamma s_{ij}^2 \nonumber\\
\quad \quad s.t. \quad s_{ij} \in (0, 1), \mathbf{I}^T s_i = 1, rank(\mathbf{L}_s) = n - c \label{CAN_EQN}
\eeqn
where $\gamma$ is the adjusted factor and $\mathbf{I}$ is denoted as the identity matrix. 
$\mathbf{L}_s$ is the Laplacian operator defined by similar matrix
$s = {s_1, s_2, \ldots, s_n}$. 
The connected relationships of a graph can be learned by implementing the rank constraint $rank(\mathbf{L}_s)$, where $n$ is equal to the number of samples and $c$ is the number of connected components of $s$.

To solve Problem \ref{CAN_EQN}, we employ the Clustering with Adaptive $K$-Nearest Neighbors Selection \cite{wang2023novel, Yuqin2024, du2016study}, which can enhance the performances in terms of effectiveness, adaptability and robustness. 
This mechanism would adapt to the non-linear and complex data structures as well as diverse data distributions in the smartgrid systems.
In particular, we utilize the non-parametric supervised learning method of k-nearest neighbors for selecting  K-nearest neighbor information of points to establish local density.
As a result, it would improve clustering accuracy and avoid the cases of misclassification.
\vspace{-0.3cm}
\section{Details of Implementation}
\vspace{-0.3cm}

\noindent \textbf{Dataset Formulation}: We perform testing in multiple IEEE bus benchmark systems, where we consider many environment conditions, such as noise, measurement errors and communication errors.
We summarize three types of environment conditions that are created by modifying measurements:

\noindent 1) Gaussian noise: The noise considered has zero mean and the standard deviation, $\sigma_{n}$ ($\sigma_{n} = 10^{-SNR/20}$), where SNR is set at 45 dB. 

\noindent 2) Data loss of buses: In manipulation, we randomly drop the data of $N_d$ buses per data sample in the test dataset (by setting the measured values to 0).

\noindent 3) Random data loss for measured data: We set each measurement at all buses by 0 with probability $P_l$.

\noindent \textbf{Algorithm Setup}: We investigate the performance of proposed intelligent diagnosis method for fault location.
Comparisons with baseline models are provided in detail.
We also visualize the hidden features of samples in the test
dataset to demonstrate that the proposed GNN model is able
to learn more robust representations from data.
In short, we present the algorithm setup as follows: 1) We firstly perform \textbf{Data acquisition}, where the raw vibration or current signals are measured under varying load and data acquisition system. 2) We perform the \textbf{Graph construction} based on \ref{CAN_TEXT}, where graph is composed of node features and neighbor connections. By using the CAN method, the neighbor relationships are constructed based on frequency spectrum of different samples. 3) Following that, we use the combined SSCRF and GAT (see \ref{GNN_GAT_CRF}) for learning node representations and modeling the label dependency.

\noindent \textbf{Dataset preparation}: We use the GNN and traditional Multi Layer Perceptron (MLP) for modeling and generating the datasets for power flows of specific power grid topology \cite{hansen2022power}.

\vspace{-10pt}
\section{Numerical Results and Discussion on Applications}
\label{Results}
\vspace{-10pt}

At our lab, we will develop the efficient computational framework for collateral damage analysis, which can be applied to secure, resilient and reliable powergrid. 
One example use-case is powergrid collateral damage analysis and fault diagnosis as well as self-healing. 
The other possible use-cases are cyber securities in smartgrid including intrusion detection, classification and event-triggered control.
In particular, the main goal here is to derive the machine learning models, which 1) can efficiently detect any possible failure and quickly locate any concerned nodes in the power network; 2) can rapidly perform load predicting task, where its results would be used for future power risk avoidance, mitigation and self-healing operation. To perform that, we need to develop the machine learning-based predicting mechanism so that we can determine the outputs of unknown voltage amplitude and voltage angle with the inputs of active power and reactive power. So, we develop the algorithms for 1) FCNN based predicting algorithm, 2) GNN based predicting algorithm, 3) GAT based predicting algorithm. 
In short, we need to predict the power flow (consisting of active power, reactive power, voltage amplitude and voltage angle), which is then used as an supporting data for these applications.
In the following, we will present the testing and validation of our proposed Dynamic Power Flow Analysis and Fault Characteristics using Graph Attention Networks, as well as make a fair comparison to the state-of-the-arts.

Firstly, we performed data generation for the power grid, which includes the dataset of power flow at the normal operation, at the failure scenarios (short-circuit nodes, cascading trips, random disruptions, etc.). 
We generated the power flow datasets for multiple bus power grids, including IEEE 14-bus feeder, IEEE 37-bus feeder, IEEE 128-bus feeder and IEEE 8500-bus feeder. 
We also generated datasets for single-phase modeling and three-phase modeling scenarios. 
GNN is used for integrating the power flow of all the nodes by leveraging the power grid connectivity information and the relationship of parameters. 
So GNN can generalize the inter-node relationship in graph data, which is used for training and testing models. 
Then, we worked on the GAT and enhanced the proposed GAT framework by utilizing some mechanisms, such as linearization and approximation. 
Furthermore, we proposed the combination of GAT and SSCRF as well as developed the CAN method for graph clustering, which enhanced the accuracy performance (for simplicity, we use term GAT for our proposal). 
We also performed the testing for the obtained dataset of power grids by using the derived algorithms.

Let take a closer look on one example as follows. 
We generated the dataset for the IEEE 14 bus power grid, which is used for training the models and for testing our developed algorithms. 
In particular, we generated 1 dataset for training, 1 dataset for validation and 100 datasets for testing. 
Each dataset is the time-series data, which consists of 2000 data points. 
To make it, we vary the load randomly with the variants of 50\% of the original load values.
We can also generate the datasets for the other IEEE bus power grids, such as IEEE 37-bus feeder, IEEE 123-bus feeder and IEEE 8500-bus feeder. 

In the following training and testing, we allow the FCNN run with maximum number of epochs of 10,000, while we limit maximum number of epochs of 2,000 for both GNN and GAT. 
We have the following observations.

\begin{figure}[!t]
\vspace{-0.5cm}
\centerline{\includegraphics[width=0.4\textwidth]{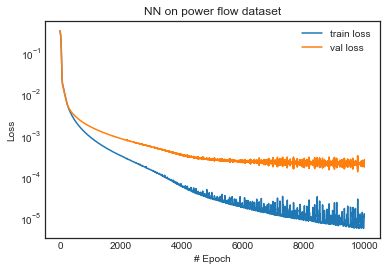}}
\vspace{-0.3cm}
	\caption{The use of FCNN with full data train size of 100\%. It would be done training with 10,000 epochs.}
	\label{Full_FCNN}
 \vspace{-0.3cm}
\end{figure}

\begin{figure}[!t]
\vspace{-0.5cm}
\centerline{\includegraphics[width=0.4\textwidth]{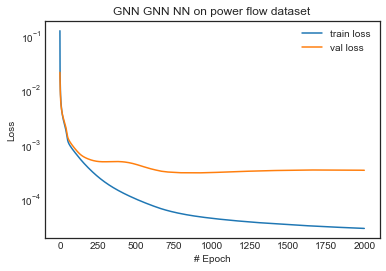}}
\vspace{-0.3cm}
	\caption{The use of GNN with full data train size of 100\%. It would be done training with 2000 epochs.}
	\label{Full_GNN2NN}
 \vspace{-0.3cm}
\end{figure}

\begin{figure}[!t]
\vspace{-0.5cm}
\centerline{\includegraphics[width=0.4\textwidth]{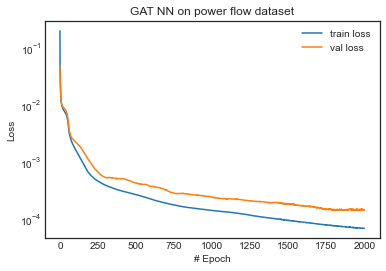}}
\vspace{-0.3cm}
	\caption{The use of GAT with lower data train size of 100\%. It would be done training with 2000 epochs..}
	\label{Full_GAT2NN}
 \vspace{-0.3cm}
\end{figure}

\begin{figure}
\vspace{-0.3cm}
    \centering
    \begin{subfigure}{.495\textwidth}
    \centering
    \includegraphics[width=.95\linewidth]{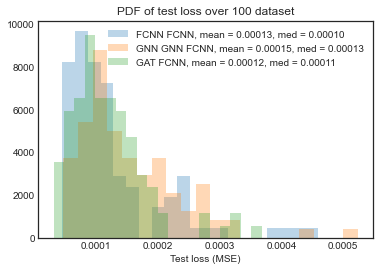}
\vspace{-0.3cm}
	\caption{Histogram of MSE performance.}
	\label{Full_hist_MSE}
    \end{subfigure}
    \hfill
    \begin{subfigure}{.48\textwidth}
    \centering
    \includegraphics[width=.95\linewidth]{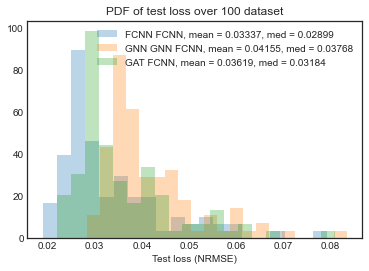}
\vspace{-0.3cm}
	\caption{Histogram of NRMSE performance.}
	\label{Full_hist_NRMSE}
    \end{subfigure}
\vspace{-0.3cm}
    \caption{Testing using GAT, GNN and FCNN with lower data train size of 100\%.}
\vspace{-0.7cm}
\end{figure}

\noindent\textbf{Observation 1) Comparison between different Machine Learning Methods.} With full dataset training, Figs. \ref{Full_FCNN}, \ref{Full_GNN2NN}, and \ref{Full_GAT2NN}  demonstrated the training loss and validation loss for FCNN, GNN and GAT respectively. 
The validation loss of GAT still decreases when the number of epochs reaches 2,000, while the validation loss of GNN is saturated at the number of epochs of 600. 
Therefore, we can see the histogram demonstrations of MSE and NRMSE for FCNN, GNN and GAT in Figs. \ref{Full_hist_MSE} and \ref{Full_hist_NRMSE} that the FCNN, GNN and GAT achieve the similar testing performance. 
It implies that when we use the full dataset for training, there is slight difference between the mechanisms. 
The need of geometric mechanisms (GNN and GAT) is only for the case of missing data in the training stage. 
So, we do investigate the case lower data training size, which is presented in the next observation.

\begin{figure}
\vspace{-0.6cm}
    \centering
    \begin{subfigure}{.48\textwidth}
    \centering
    \includegraphics[width=.95\linewidth]{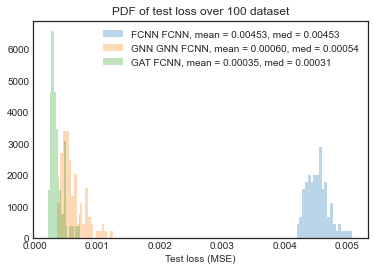}
\vspace{-0.3cm}
	\caption{Histogram of MSE performance.}
	\label{20_hist_MSE}
    \end{subfigure}
    \hfill
    \begin{subfigure}{.48\textwidth}
    \centering
    \includegraphics[width=.95\linewidth]{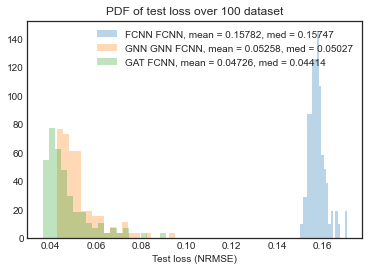}
\vspace{-0.3cm}
	\caption{Histogram of NRMSE performance.}
	\label{20_hist_NRMSE}
    \end{subfigure}
\vspace{-0.3cm}
    \caption{Testing using GAT, GNN and FCNN with lower data train size of 20\%.}
\vspace{-0.3cm}
\end{figure}

\begin{figure}[!t]
\vspace{-0.5cm}
\centerline{\includegraphics[width=0.4\textwidth]{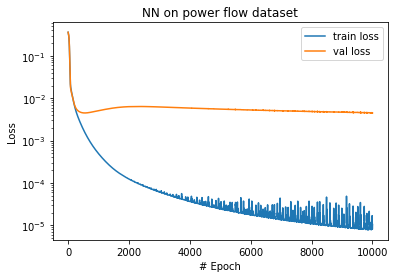}}
\vspace{-0.3cm}
	\caption{The use of FCNN with lower data train size of 20\%. It would be done training with 10,000 epochs.}
	\label{20_FCNN}
 \vspace{-0.3cm}
\end{figure}

\begin{figure}[!t]
\centerline{\includegraphics[width=0.4\textwidth]{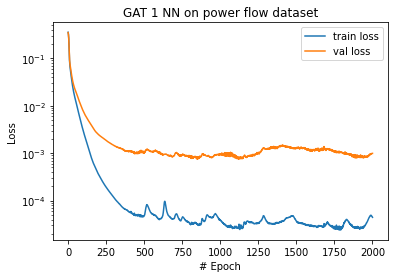}}
\vspace{-0.3cm}
	\caption{The use of GAT with lower data train size of 20\% and the ReLu  activation function.}
	\label{20_GAT_ReLu}
 \vspace{-0.3cm}
\end{figure}

\begin{figure}[!t]
\vspace{-0.3cm}
\centerline{\includegraphics[width=0.4\textwidth]{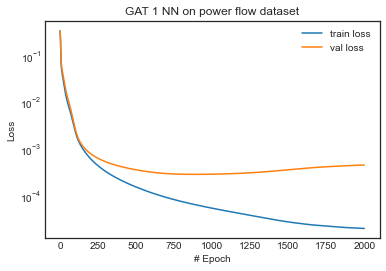}}
\vspace{-0.3cm}
	\caption{The use of GAT with lower data train size of 20\% and the Tanh activation function.}
	\label{20_GAT_Tanh}
 \vspace{-0.5cm}
\end{figure}

\noindent\textbf{Observation 2) Consideration for the Scenario of Missing Data.} We now considered the lower training dataset, where only 20\% dataset is used for training. When we perform testing, the histogram demonstrations of MSE and NRMSE for FCNN, GNN and GAT in Figs. \ref{20_hist_MSE} and \ref{20_hist_NRMSE} show that there are the big gaps between the region of GNN and FCNN as well as between the region of GAT and FCNN. 
Also, the GAT region shifts toward zero comparing the GNN region. We also illustrate the FCNN's performance and GAT's performance in Figs. \ref{20_FCNN}, \ref{20_GAT_ReLu} and \ref{20_GAT_Tanh}, the FCNN's performance curve of validation loss is saturated at the number of epochs of 1,500 and its saturated performance is higher than the performance of GNN and GAT. 
Here, the FCNN's performance curve of validation loss is saturated at the number of epochs of 1,500 and its saturated loss performance is higher than the performance of GAT.



\noindent\textbf{Observation 3) Performance Comparison when Considering Different Activation Functions.} There are many activation functions for training, such as the Sigmoid, Tanh and Rectified Linear Unit (ReLu) functions. 
However, we choose the Tanh function because it has the better advantages, which are presented as follows.
In the use of Sigmoid function, the gradient updates go too far in different directions. 
This phenomenon is called the problem of vanishing gradients. 
It is too hard to optimize the model with the output in the range of [0, 1]. 
So, the use of Sigmoid function will increase the computational time.
Given this observation, we only test this activation function for the FCNN in Fig. \ref{20_FCNN}. 
The other ReLu activation function is used some application with the advantages of avoiding and rectifying vanishing gradient problem as well as having less computationally expensive than Sigmoid. 
However, there are some cases that some gradients can be fragile during training and can die. 
Of course, we obtain dead neurons. 
To make a clearer explanation, let us consider the activations in the region ($x<0$) of ReLu function. 
In this region, the gradient is 0 because we cannot regulate the weights during descent. 
It means that those neurons jumping to that state will not respond to any inputs and feedback errors.
The Tanh function can help to solve the nonzero centered problem of the sigmoid function. 
Also, the Tanh function will compress the real-valued data to the range of [-1, 1]. 
Furthermore, the Tanh function is non-linear, continuously differentiable, monotonic and has a fixed output range. 
Therefore, it is simple and is good for classifier. 
We can observe the problem in Figs. \ref{20_FCNN}, \ref{20_GAT_ReLu} and \ref{20_GAT_Tanh}.

\vspace{-0.3cm}
\section{Conclusion and Future Directions}
\vspace{-0.3cm}
\subsection{Conclusion}
We have deep investigation of ML-based schemes in various applications within the concept of the powergrid, i.e. the fault diagnosis and self-healing techniques and cyber-physical security. 
We derive the joint GAT, CAN and SSCRF framework for power flow analysis, which can be used for further processing tasks, such as fault diagnosis, damage analysis, self-healing and condition monitoring.
Extensive numerical results have been presented to demonstrate the significant gains of our joint GAT, CAN and SSCRF framework to the accuracy and CPU time performances of power flow prediction.
It implies that our proposed framework can achieve the computational efficiency and would be applied to the large scale network. 
In the context of cyber resilience, this mechanism can support for big data management (not required labels to all the nodes and their data) and full utilization of the relationship/dependency between network elements as well as achieve the high efficiency and effectiveness.
\vspace{-0.5cm}
\subsection{Future Directions}
\vspace{-0.1cm}

Although ML-based schemes have exhibited great potential towards realizing a resilient, reliable, and secure powergrid, there are highlighted areas/directions that need improvements. \\
1) Data augmentation: The ML-based fault diagnostic schemes highly depend on reliability of the data. Therefore, we need to check the fitness of the dataset. So we would develop the consistent and definite standards for verification of reliability of the data. \\
2) We will apply our proposed framework to different use-case application. One example is that our scalable ML-based fast control and optimization solutions would be investigated and adapted with fast classification feature to distributed energy resources (DERs) in centralized or islanded power grid. This is essential to overcome the existing computational burdens considering communication delays. Specifically, we need to develop autonomous self-learning methods (such as adjusting the active power delivered by DERs), which would combine with DERs’ generation prediction and grid inertia estimation to prevent the potential deleterious voltage/frequency sags and swells. \\
3) We will apply our schemes to ML-based IDSs, which can address the model-based IDS challenges. In this context, we will evaluate our proposed mechanism by using real data and real-life scenarios. Recall that the main problem of IDS is that some attack models are rare and extreme and hence, it is impossible to acquire all data of abnormal behavior. Furthermore, the power grid system highly depends on environmental conditions such that it makes change gradually on determining what observation is normal or abnormal. It means that the IDSs are required to feature quick reaction to misbehaving grid clusters due to the system and environment and adversarial manipulations.

\vspace{10pt}
\noindent\textbf{Acknowledgment}: 
This work was supported in part by the Commonwealth Cyber Initiative (CCI) Experiential Learning Program and the South Big Data Innovation Hub Partnership Nucleation program.

\bibliography{references}
\bibliographystyle{IEEEtran}

\end{document}